\documentclass[final,3p,times]{elsarticle}

\usepackage{amssymb}
\usepackage{bm}
\usepackage{amsmath}
\usepackage[ruled]{algorithm2e}
\usepackage[usenames,dvipsnames]{color}
\usepackage{array}
\usepackage{subcaption}
\usepackage{graphicx}
\usepackage{lineno}
\usepackage{tikz}
\usetikzlibrary{shapes, arrows.meta, positioning, calc}
\usepackage{booktabs}
\usepackage{float}
\usepackage[labelfont=bf,textfont=it]{caption}
\usepackage{placeins}
\usepackage{longtable}
\usepackage{enumitem}

\usepackage{hyperref}
\hypersetup{
    colorlinks,
    linkcolor={magenta},
    citecolor={blue},
    urlcolor={blue!80!black},
    breaklinks=true,
    plainpages=true
}

\newcommand{\cref}[3]{\hyperref[#2]{#1~\ref*{#2}{#3}}}
\newcommand{\crefs}[3]{\hyperref[#2]{#1~\ref*{#2}-\ref*{#3}}}
\newcommand{\colref}[2]{\hyperref[#2]{#1~\ref*{#2}}}

\newcommand{\figref}[1]{\colref{Figure}{#1}}
\newcommand{\subfigref}[2]{\cref{Figure}{#1}{#2}}
\newcommand{\figsref}[2]{\crefs{Figures}{#1}{#2}}
\newcommand{\secref}[1]{\colref{Section}{#1}}

\biboptions{sort&compress}
\newcommand{\doi}[1]{\textsc{doi}: \href{http://dx.doi.org/#1}{\nolinkurl{#1}}}

\begin{document}

\journal{Computers and Electronics in Agriculture}

\begin{frontmatter}


\title{Accessing the Effect of Phyllotaxy and Planting Density on Light Use Efficiency in Field-Grown Maize using 3D Reconstructions}

\author[inst1]{Nasla Saleem}
\author[inst1]{Talukder Zaki Jubery}
\author[inst1]{Aditya Balu}
\author[inst2]{Yan  Zhou}
\author[inst2]{Yawei Li}
\author[inst2]{Patrick S. Schnable}
\author[inst1]{Adarsh Krishnamurthy}
\author[inst2]{Baskar Ganapathysubramanian\texorpdfstring{\corref{cor1}}{}}

\affiliation[inst1]{organization={Department of Mechanical Engineering},
            addressline={Iowa State University}, 
            city={Ames},
            postcode={50011}, 
            state={Iowa},
            country={USA}}

\affiliation[inst2]{organization={Department of Agronomy},
            addressline={Iowa State University}, 
            city={Ames},
            postcode={50011}, 
            state={Iowa},
            country={USA}}

\cortext[cor1]{Corresponding Authors}
\begin{abstract}
High-density planting is a widely adopted strategy to enhance maize productivity, yet it introduces challenges such as increased interplant competition and shading, which can limit light capture and overall yield potential. In response, some maize plants naturally reorient their canopies to optimize light capture, a process known as canopy reorientation. Understanding this adaptive response and its impact on light capture is crucial for maximizing agricultural yield potential. This study introduces an end-to-end framework that integrates realistic 3D reconstructions of field-grown maize with photosynthetically active radiation (PAR) modeling to assess the effects of phyllotaxy and planting density on light interception. In particular, using 3D point clouds derived from field data, virtual fields for a diverse set of maize genotypes were constructed and validated against field PAR measurements. Using this framework, we present detailed analyses of the impact of canopy orientations, plant and row spacings, and planting row directions on PAR interception throughout a typical growing season. Our findings highlight significant variations in light interception efficiency across different planting densities and canopy orientations. By elucidating the relationship between canopy architecture and light capture, this study offers valuable guidance for optimizing maize breeding and cultivation strategies across diverse agricultural settings.
\end{abstract}

\begin{graphicalabstract}
\includegraphics[width=\linewidth, trim=10 310 10 275, clip]{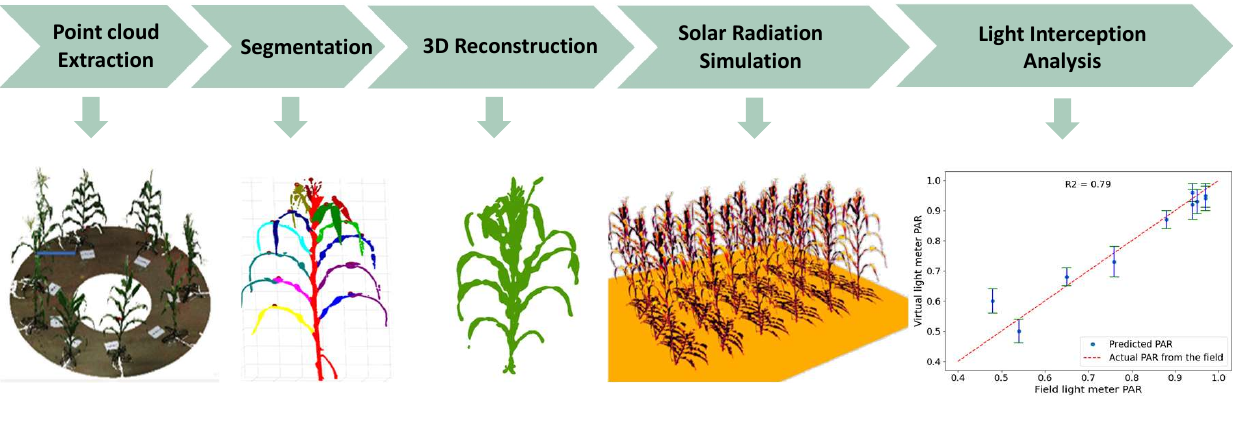}
\end{graphicalabstract}

\begin{highlights}
\item Virtual measurement framework for photosynthetically active radiation (PAR) of field maize.
\item Framework validated with actual field measurements.
\item Explored PAR interception under varying maize leaf azimuth angles and canopy reorientation.
\item Analyzed the impact of planting row directions on PAR interception.
\item Investigated the effects of planting densities on PAR interception.
\end{highlights}


\end{frontmatter}



\section{Introduction}

\begin{figure}[t!]
    \centering
    \includegraphics[width=\linewidth, trim=10 310 10 275, clip]{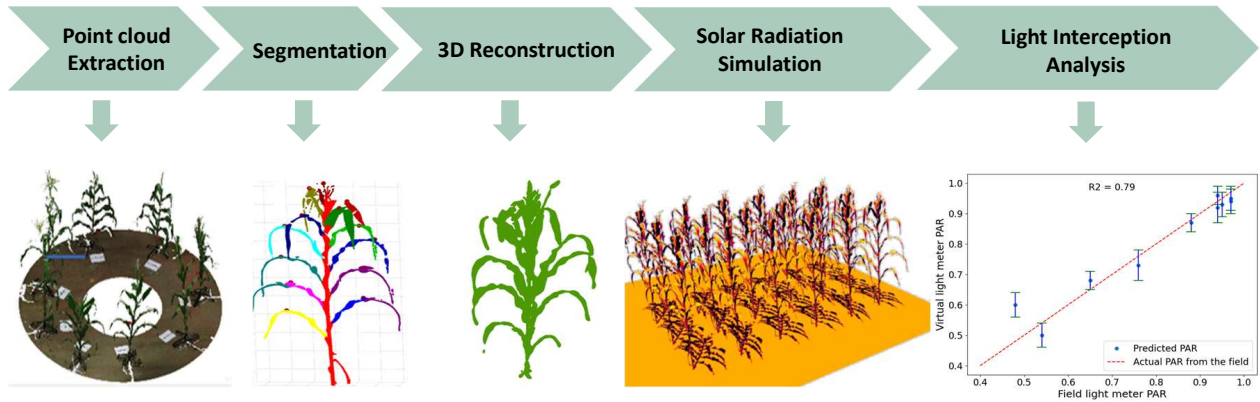}
    \caption{Overview of the framework for PAR interception analysis:  
    (1) Point Cloud Extraction: Point clouds of field-grown maize were captured using a LiDAR scanner.
    (2) Segmentation: Different plant organs were segmented from individual plants.  
    (3) 3D Reconstruction: Segmented organs were reconstructed separately to create accurate 3D plant models.  
    (4) Solar Radiation Simulation: Plants of each genotype were replicated to form a virtual field, and solar radiation was simulated using Helios software.  
    (5) Light Interception Analysis: Analyzing PAR under varying planting densities, row directions, and leaf azimuth arrangements.}
    \label{fig: framework}
\end{figure}

 Maize (Zea mays L.) stands as a cornerstone crop, pivotal to both the U.S. economy and global agriculture~\citep{erenstein2022global}. With an annual production of 1.2 billion tons, maize is the most important cereal crop worldwide~\citep{canton2021food}. Achieving high maize yields is crucial to meet the escalating demands for food and industrial products. Over the past several decades, the adoption of higher planting densities has significantly boosted maize grain yields~\citep{tokatlidis2004review, duvick2004long}. This has been enabled by the enhanced tolerance of modern maize hybrids to increased planting densities~\citep{berzsenyi2012density, gonzalez2018maize}.
 
 Recent studies suggest that while increased planting densities may yield less grain per plant, modern hybrids can achieve greater overall yields as a result of the improved tolerance to higher planting densities, which result in more ears per unit land area~\citep{assefa2018analysis, duvick1999post, duvick2005genetic}. However, increased shading from adjacent plants at high densities can still cause significant yield losses~\citep{earley1966effect, reed1988shading}. Under conditions where water,  nitrogen and ther nutrients are not limiting, the relationship between planting density and yield is primarily influenced by the plant's ability to manage interplant competition while maximizing light interception. Thus, canopy architecture plays a pivotal role in determining the amount of shade received by adjacent plants, affecting light penetration and ultimately influencing per plant yields~\citep{lambert1978leaf, toler1999corn, hammer2009can}.
 
 Maximizing light interception in densely planted maize fields is essential, as interplant competition can lead to significant yield losses due to shading~\citep{tokatlidis2004review, al2015maize,sangoi2001understanding}. The efficiency of light interception in plants is usually calculated by the fraction of photosynthetically active radiation (PAR) -- defined as light within the 400 to 700 nm spectral range -- intercepted by the canopy~\citep{wimalasekera2019effect}. As mentioned earlier, canopy architecture, which includes features such as leaf area, leaf angles,  and azimuthal leaf orientations, plays a critical role in determining light distribution within the canopy~\citep{niinemets2010review}. Optimizing canopy photosynthetic capacity during the flowering stage,  when the kernel set is being determined, is crucial for minimizing the negative effects of interplant competition. Thus, understanding the role of canopy architecture in crop yield dynamics is essential, particularly in densely planted settings, to provide insights into optimizing yield outcomes through strategic management of planting density and canopy architcture~\citep{zhu2012elements, hammer2002future}. However, the architectural characteristics of maize canopy remain not well characterized, primarily due to the lack of powerful tools capable of testing the myriad combinations of architectural and eco-physiological traits~\citep{zhu2012elements, rotter2015use}. In addition, most studies were conducted based on a single inbred or hybrid, which could potentially lead to inaccurate predictions when the resulting models are applied to other hybrids with different genetic backgrounds\citep{rotter2015use}.
 
 To mitigate the effects of high-density planting, some maize genotypes  possess the capacity to adapt their architectures to altered planting distribution patterns and densities. One of the notable mechanisms is leaf reorientation, where maize plants can alter the azimuthal orientations of their leaves in response to interplant competition~\citep{maddonni2002maize, drouet1997spatial}. Changes in the leaf azimuth due to increased planting density have been documented in previous studies, showing that under highly rectangular distribution patterns (where the distance between rows is much greater than the distance between plants in the same row), maize plants can orient their leaves perpendicular to the row~\citep{toler1999corn, drouet2008does, maddonni2001light}. \citet{zhou2024genetic} recently reported the genetic mechanism by which some maize genotypes to alter the azimuthal orientations of their leaves during development in coordination with adjacent plants. This adaptation enhances light interception by decreasing mutual shading. The nature of the responsible genes and the density-dependent nature of this re-orientation demonstrate that the process is a response to shade avoidance. Therefore, it is essential to understand how these canopy reorientations and different planting densities impact light interception and ultimately influence crop yield.

 Measuring the direct impact of canopy architecture on light interception is challenging due to the spatiotemporal variability in light distribution. Functional–structural plant models (FSPMs) have been developed and widely used to simulate plant architecture and physiological functions under various environmental conditions~\citep{drouet2003graal, song2008analysis}. However, the lack of detailed descriptions at the organ level in existing FSPMs can limit the accuracy of assessing actual light interception~\citep{kim2020use}. Recent advances in sensor-based technologies, including LiDAR, depth cameras, time-of-flight cameras, and multi-view stereo-derived point clouds from images, have facilitated the creation of more realistic three-dimensional plant canopy models~\citep{paturkar2021making}. When integrated with radiative transfer models, these technologies significantly enhance our understanding of how light interacts with plant structures in three-dimensional spaces~\citep{okura20223d}. 
 
 Several studies have contributed to this field, including research on canopy photosynthesis models that utilize 3D modeling to refine predictions of light utilization, highlighting the necessity of realistic structures for accurate simulations~\citep{gu2022use, xiao2023importance}. Additionally, investigations into light interception characteristics across different plant species have underscored the role of canopy architecture in optimizing light capture~\citep{sultana2023competition, zhao2024fine}. In maize, \citet{song2023quantifying} proposed a modeling pipeline to quantify light interception in intercropped canopies and enhance our understanding of canopy photosynthesis. Similarly, \citet{zhu2020quantification} proposed a novel approach that emphasizes the importance of accurate light interception assessments in different maize varieties. Nevertheless, the effects of canopy reorientations and varying row directions in high-density planting on light interception remain underexplored, making it crucial to understand their impact on specific genotypes in breeding programs, as well as the effects of planting density and row orientations for effective management practices.
 
 To address these gaps, we have designed an end-to-end framework—from field reconstruction to computational evaluation of PAR—to accurately measure PAR under varying canopy conditions. Our objectives are twofold: first, to establish a comprehensive 3D canopy modeling framework for precise PAR measurement, and second, to conduct a detailed analysis of the effects of canopy reorientation, planting density, row orientations, and plant spacing on canopy light interception.

Our specific contributions are:
\begin{itemize}[itemsep=0pt,topsep=0pt]
\item Development and validation of a virtual framework to accurately measure  PAR interception by maize canopies using LiDAR and solar radiation simulation.
\item Exploration of PAR interception dynamics under varying patterns of maize leaf azimuth arrangement (typically causing canopy reorientation) throughout a day and throughout the growing season.
\item Investigation into the impacts of different row directions and planting densities, including changes in inter-plant and inter-row distances, on PAR interception by maize plants.
\end{itemize}

The remainder of the paper is organized as follows. In \secref{sec:Methods}, we provide a detailed exploration of each step in the end-to-end framework for PAR interception analysis, including data collection (\secref{subsec:data}), point cloud extraction (\secref{subsec:3d}), segmentation and 3D reconstruction, virtual field creation (\secref{subsec:field}), and solar radiation simulation for analyzing light interception dynamics (\secref{subsec:helios}). In \secref{sec:results}, we present the validation of our framework and results of the impact of different planting scenarios in PAR interception. The paper concludes with a summary and discussion of future research directions in \secref{sec:conclusion}.

\section{Materials and Methods}\label{sec:Methods}

Our approach (\figref{fig: framework}) includes an end-to-end framework for analyzing PAR interception in maize canopies. This framework involves capturing point clouds of field-grown maize using a LiDAR scanner, segmenting plant organs, reconstructing accurate 3D models, simulating solar radiation in a virtual field environment, and analyzing PAR interception under various planting densities, row directions, and leaf azimuth arrangements. In the following sections, we describe each step of our framework in detail.

 \begin{figure}[t!]
    \centering
    \includegraphics[width=0.94\linewidth, trim=0 3 0 3, clip]{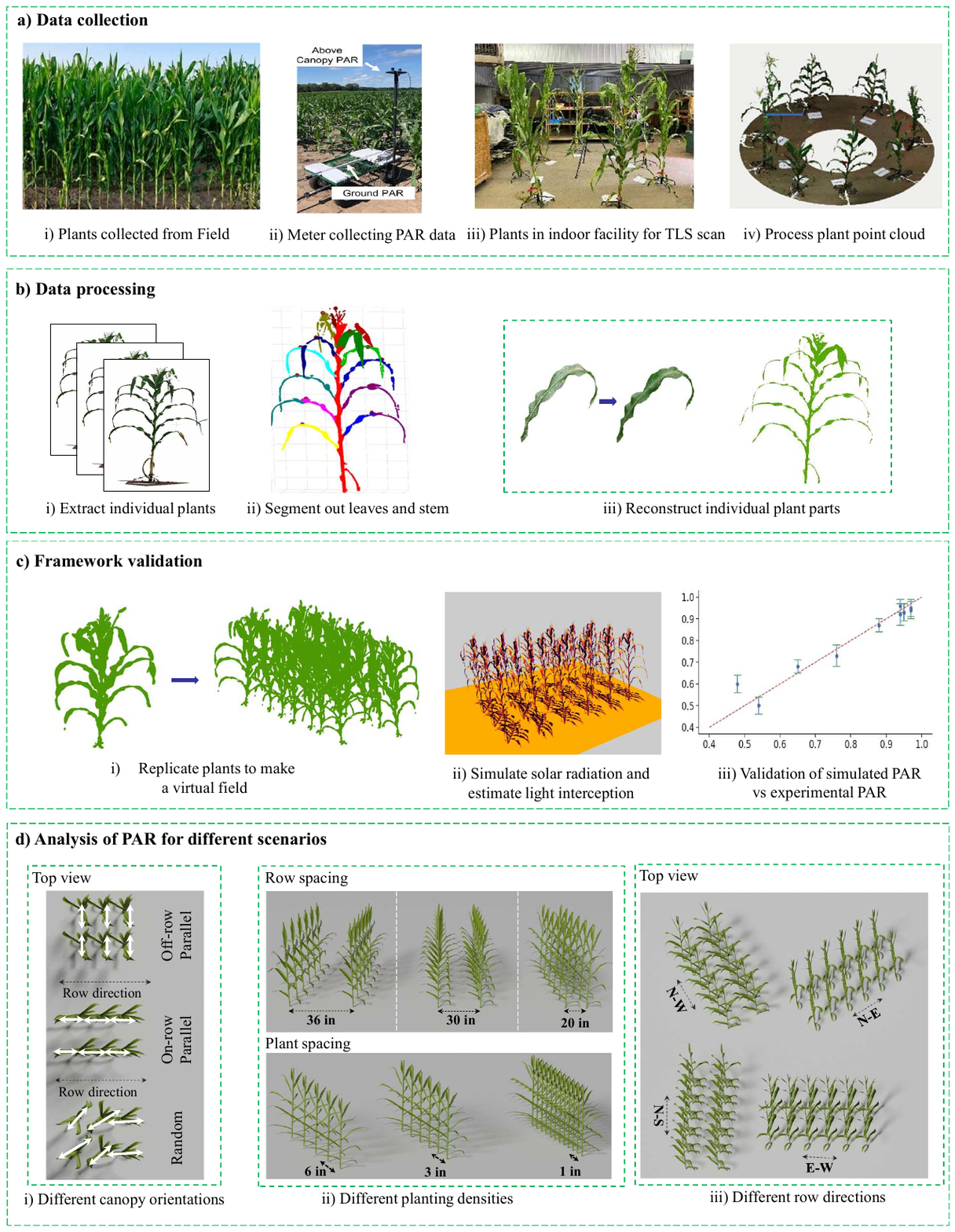}
    \caption{Detailed breakdown of the pipeline for 3D canopy modeling and PAR analysis, including (A) data collection of plant materials and PAR data from the field, (B) data processing, (C) framework validation with field PAR measurements, and (D) analyzing PAR for different scenarios.}
    \label{fig: methods}
\end{figure}

\subsection{Plant materials and PAR data collection}
\label{subsec:data}

We used 10 maize (Zea mays L.) inbred lines sourced from the  maize shoot apical meristem (SAM) diversity panel~\citep{leiboff2015genetic} for this study. These lines were planted in 4-row plots at the Iowa State University Woodruff Farm in Ames, Iowa, in the year 2021. Each plot measured 3.04 meters in row length, with plants spaced approximately 15.24 cm apart within rows and 76.2 cm apart between rows, resulting in a planting density of around 84,000 plants per hectare. All plants were collected from the field at the time of anthesis and immediately brought to indoor environment for imaging and scanning.

The fraction of intercepted PAR values of these lines was also collected and used to validate our framework. The details of data collection have been previously described by Zhou et al.~\citep{zhou2024genetic}. In brief, the PAR data used in this study were collected on August 7, 2020, 60 days after planting, when the plants in the row were at anthesis. The fraction of PAR intercepted was estimated by subtracting the ground-level PAR measurement from the average of the above-canopy measurements and then dividing this difference by the average of the above-canopy measurements.

\subsection{Acquiring and processing 3D plant LiDAR data}
\label{subsec:3d}
The field-grown plants were harvested at around the time of anthesis and transported to an indoor facility for detailed 3D imaging. High-resolution 3D point cloud data of the maize plants were captured using a Faro Focus S350 Scanner. The indoor scanning process involves arranging the plants in a circular pattern around the scanner to ensure comprehensive coverage and maximum visibility within the scanner’s line of sight, thereby minimizing overlaps. High-resolution LiDAR-TLS 3D point cloud data of the plants were acquired using the Faro Focus S350 Scanner across all 10 genotypes. The LiDAR-TLS data were processed using  FARO SCENE software~\citep{faro_scene}, including registering, refining, and denoising the acquired pointcloud for detailed analysis. The resulting point cloud dataset contains X-Y-Z coordinates, RGB color information, and normal vectors representing surfaces such as leaves and stems \subfigref{fig: methods}{.(a)}. Individual plant point clouds were then extracted and saved as separate files using the CloudCompare software~\citep{cloudcompare}.

\subsection{Segmenting and Reconstructing Individual Plant Parts}\label{subsec:seg}
Once the individual plant point clouds are acquired, they are segmented to isolate plant organs, such as leaves, stems, and other relevant parts, for accurate 3D reconstruction and analysis. The processed point clouds are segmented using the label3dmaize software~\citep{miao2021label3dmaize}, which is specifically designed for maize plant structure analysis. This software employs machine learning and geometric algorithms to detect and classify plant organs within the point cloud data. By accurately labeling plant structures, label3dmaize enables the isolation of specific parts while excluding extraneous regions, such as below-ground sections or background noise \subfigref{fig: methods}{.b).ii)}. We manually verify the quality of the segmentation.

\begin{figure}[h!]
    \centering
    \includegraphics[width=\linewidth, height=0.3\textheight, trim={10mm 100mm 10mm 100mm},clip]{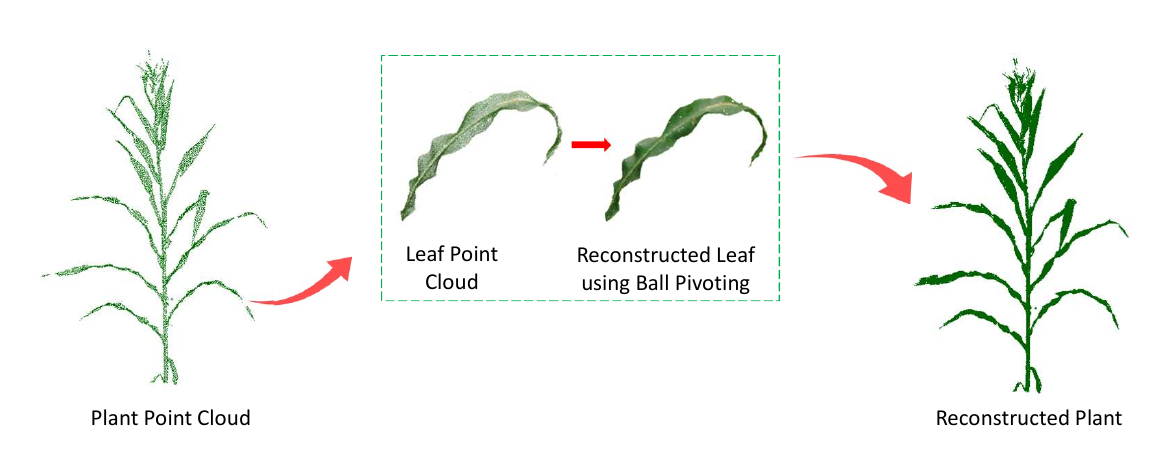}
    \caption{3D reconstruction of maize plants from point cloud data, showcasing the transformation from raw point clouds to fully reconstructed plant structures, including leaves and stems, using the ball pivoting algorithm.
    }
    \label{fig:reconstruction}
\end{figure}

Once segmented, each plant organ is prepared for 3D reconstruction. The ball pivoting algorithm~\citep{bernardini1999ball}, implemented in MeshLab~\citep{LocalChapterEvents:ItalChap:ItalianChapConf2008:129-136}, is used to generate detailed 3D meshes of segmented parts \figref{fig:reconstruction}. The ball pivoting algorithm is a surface reconstruction technique that works by rolling a virtual ball of a specified radius over the point cloud data. When the ball contacts three points simultaneously, it forms a triangular facet, which is added to the mesh. This process iteratively continues across the entire point cloud, connecting points and creating a network of triangles that represent the surface of the plant part. Key parameters, such as the ball radius, influence the granularity and smoothness of the resulting mesh. A smaller ball radius can capture fine details but may increase computational complexity, while a larger radius results in smoother surfaces at the cost of finer structural nuances. Careful parameter tuning is usually necessary to balance detail and computational efficiency. We use a ball radius of 0.5 mm.

Following the reconstruction of individual plant organs, these parts are merged to form a complete 3D plant mesh. This merging step involves aligning and combining the reconstructed meshes while ensuring seamless transitions between connected parts. The final 3D models preserve structural fidelity and provide detailed representations of the plants' morphology, enabling comprehensive phenotypic analysis and downstream applications \subfigref{fig: methods}{.b).iii)}.

\subsection{Generating Virtual Maize Plots with Relevant Field Parameters}  
\label{subsec:field}  

After reconstructing each of the 10 genotypes, virtual fields are created to simulate realistic agricultural conditions and investigate the effects of planting configurations \subfigref{fig: methods}{.c).i)}. For each genotype, virtual fields consist of 15 plants per row were created and arranged to replicate real-world planting conditions. The virtual fields are initially configured with a row spacing of 30 inches and an intra-row plant spacing of 6 inches, mimicking field conditions. To explore the influence of planting density on light interception and plant performance, the initial row and plant spacing values are subsequently systematically adjusted.  

The virtual fields incorporate several configurations to evaluate their impact on light interception and plant performance:  

\begin{itemize}  
    \item \textbf{Canopy Orientations} \subfigref{fig: methods}{.d).i)}: Leaf azimuth orientations are simulated in three configurations—on-row parallel, off-row parallel, and random orientations—to study their effects on light interception dynamics.  
    \item \textbf{Planting Densities} \subfigref{fig: methods}{.d).ii)}: Different row and plant spacings are simulated to analyze the impact of planting density on interplant competition and resource use.  
    \item \textbf{Planting Row Directions} \subfigref{fig: methods}{.d).iii)}: Rows are oriented in different directions (e.g., north-south, east-west, diagonal) to examine how planting alignment impacts light capture throughout the day.  
\end{itemize}  

These configurations and their impacts are explained in additional detail in the results section.  

\subsection{Simulating Solar Radiation on Virtual Maize Fields and Analyzing Light Interception Dynamics}\label{subsec:helios}
Solar radiation patterns were simulated on virtual maize fields using the \texttt{Helios 3D} modeling framework~\citep{bailey2019helios}, which incorporates the radiative transfer model from Bailey et al.~\citep{bailey2018reverse} (\subfigref{fig: methods}{.c).ii)}). \texttt{Helios} uses a reverse ray tracing method to compute PAR in the 400-700 nm wavelength range. In ray tracing, light rays are traced as they travel through a scene, interacting with various surfaces. Reverse ray tracing, as implemented in \texttt{Helios}, traces rays backward from the receiver (the plant surfaces) to the light source (the Sun). This method is computationally efficient, especially for complex 3D structures like plant canopies, because it only needs to trace rays that will actually interact with the surfaces.

The reverse ray tracing method is mathematically described by tracing rays backward from the plant surface to the light source. A ray in 3D space can be represented as:
\begin{equation}
\mathbf{r}(s) = \mathbf{r}_0 + s \cdot \mathbf{d}    
\end{equation}
where \( \mathbf{r}(s) \) represents the position of the ray at parameter \( s \), \( \mathbf{r}_0 \) is the starting point of the ray on the plant surface, and \( \mathbf{d} \) is the direction vector determined by the geometry of the plant and the position of the light source (the Sun).  

Once the rays are traced back to the plant surface, the amount of PAR intercepted by the plant is computed as:  
\begin{equation}
P_{\text{intercepted}} = \int_{\text{plant surface}} I_{\text{incident}} \cdot (1 - \rho - T) \, dA
\end{equation}
where \( P_{\text{intercepted}} \) denotes the intercepted PAR, \( I_{\text{incident}} \) is the incident solar radiation flux (irradiance), \( \rho \) and \( T \) represent the leaf reflectance and transmittance coefficients, respectively, and \( dA \) is an infinitesimal surface area element on the plant canopy.  

The total intercepted PAR over the daylight period, from 07:00 to 20:00, is given by:  

\begin{equation}
P_{\text{total}} = \int_{t_{begin} = 07:00}^{t_{end} = 20:00} P_{\text{intercepted}}(t) \, dt    
\end{equation}
where \( P_{\text{total}} \) is the cumulative intercepted PAR throughout the day, \( P_{\text{intercepted}}(t) \) denotes the time-dependent intercepted PAR at time \( t \).



Solar radiation is simulated at each time step within the virtual maize fields based on the radiative transfer model. These simulated fields, consisting of 3D canopy models of maize plants, were used to examine the interaction of light with the canopies and quantify the amount of PAR intercepted by the plants. For the simulations, the leaf reflectance and transmittance coefficients were set to 0.1, which reflect typical maize leaf behavior in terms of light reflection and transmission~\citep{earl1997maize}. To ensure accurate modeling of light scattering and interactions with the canopy, five scattering iterations were used. A periodic boundary condition was applied to reduce edge effects, making the simulated canopy behave as part of a repeating field. Simulations were conducted for three different locations—Ames, Iowa; Thomas County, Kansas; and Bismarck, North Dakota—during the period from July 15 to August 15. Detailed simulation parameters, including environmental settings, are provided in \tablename~\ref{table:supplementary_input_parameters}.

\section{Results}\label{sec:results}

\subsection{Validation of Simulated PAR Values Against Field Measurements}
To ensure the accuracy of our simulation framework, we designed a validation experiment by replicating the same field measurement setup in our simulations. This involved matching the date, time, and location of field measurements for consistency. Simulated PAR values were recorded using a virtual PAR meter and compared with field-measured PAR values for all 10 genotypes.

\begin{figure}[b!]
    \centering
    \includegraphics[width=\linewidth, height=0.3\textheight, trim={20mm 65mm 20mm 60mm},clip]{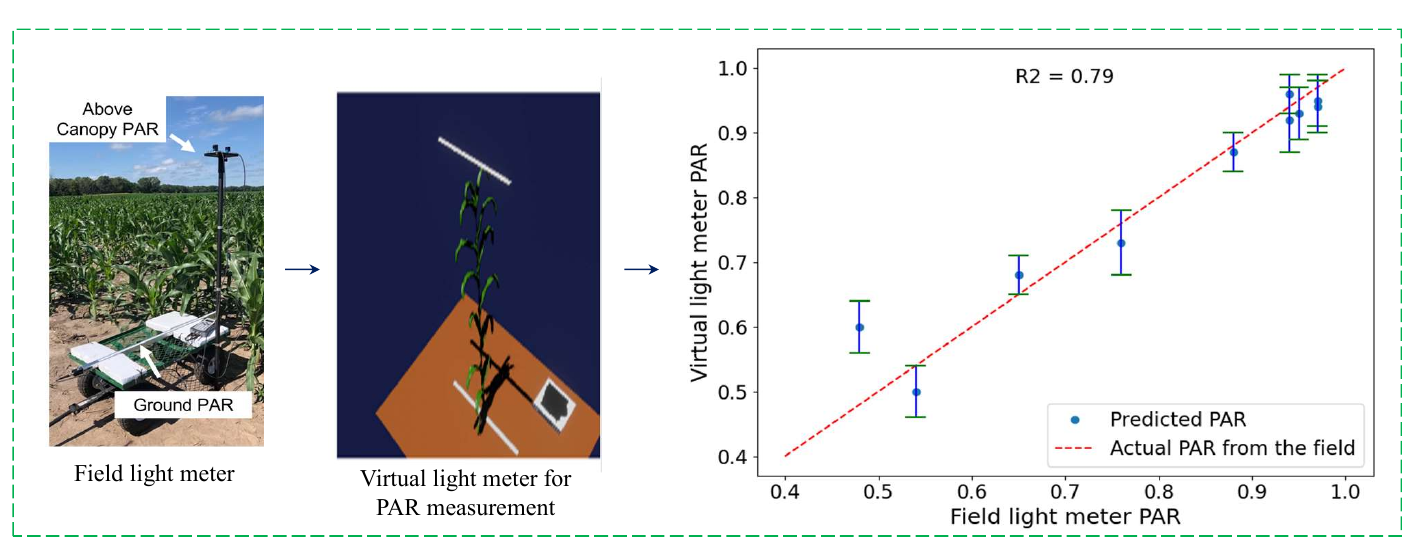}
    \caption{Validation of framework accuracy: Comparison of simulated PAR interception from a virtual light meter with field measurements of PAR interception recorded at the same date and time.}
    \label{fig:PAR validation}
\end{figure}

The fraction of PAR intercepted by the canopies of the ten genotypes, based on field measurements, ranged from 0.42 to 0.96, capturing the variation in light interception across the different genotypes. When we compared the simulated and field-measured PAR values, the results showed a strong correlation, with a coefficient of determination (R²) of 0.79 \figref{fig:PAR validation}. These results confirm that our simulations closely align with real-world measurements. This validation gives us confidence in applying the framework to explore more complex planting configurations and their impact on light interception.

\begin{figure}[t!]
\centering
\begin{minipage}{0.48\linewidth}
    \centering
    \includegraphics[width=0.99\linewidth, height=0.6\textheight, trim={60mm 70mm 75mm 70mm},clip]{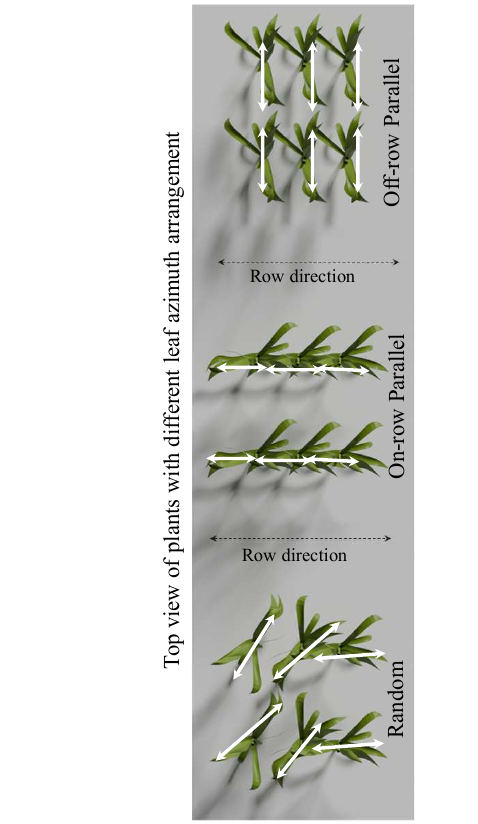}
\end{minipage}%
\begin{minipage}{0.5\linewidth}
    \centering
    \includegraphics[width=0.99\linewidth, height=0.3\textheight, trim={0mm 0mm 0mm 0mm},clip]{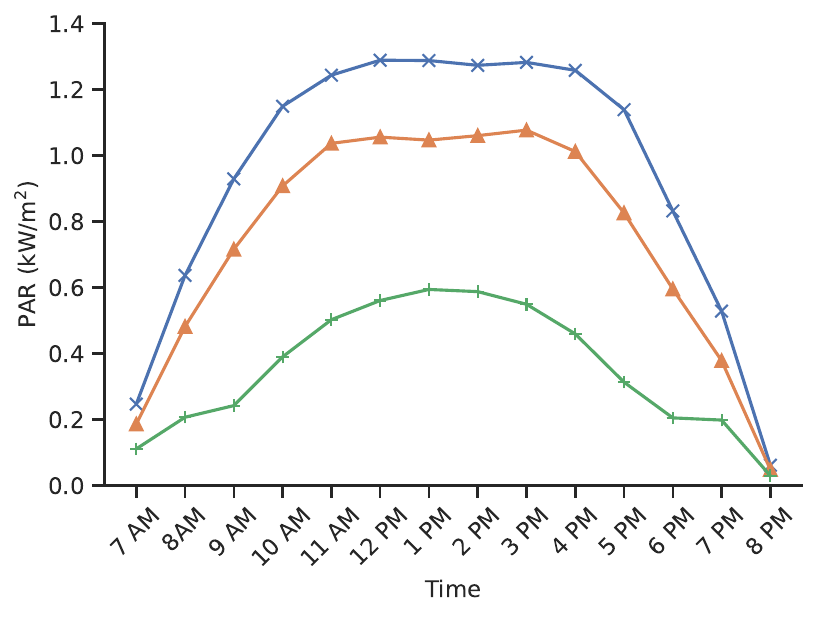}
    \includegraphics[width=0.99\linewidth, height=0.3\textheight, trim={0mm 0mm 0mm 0mm},clip]{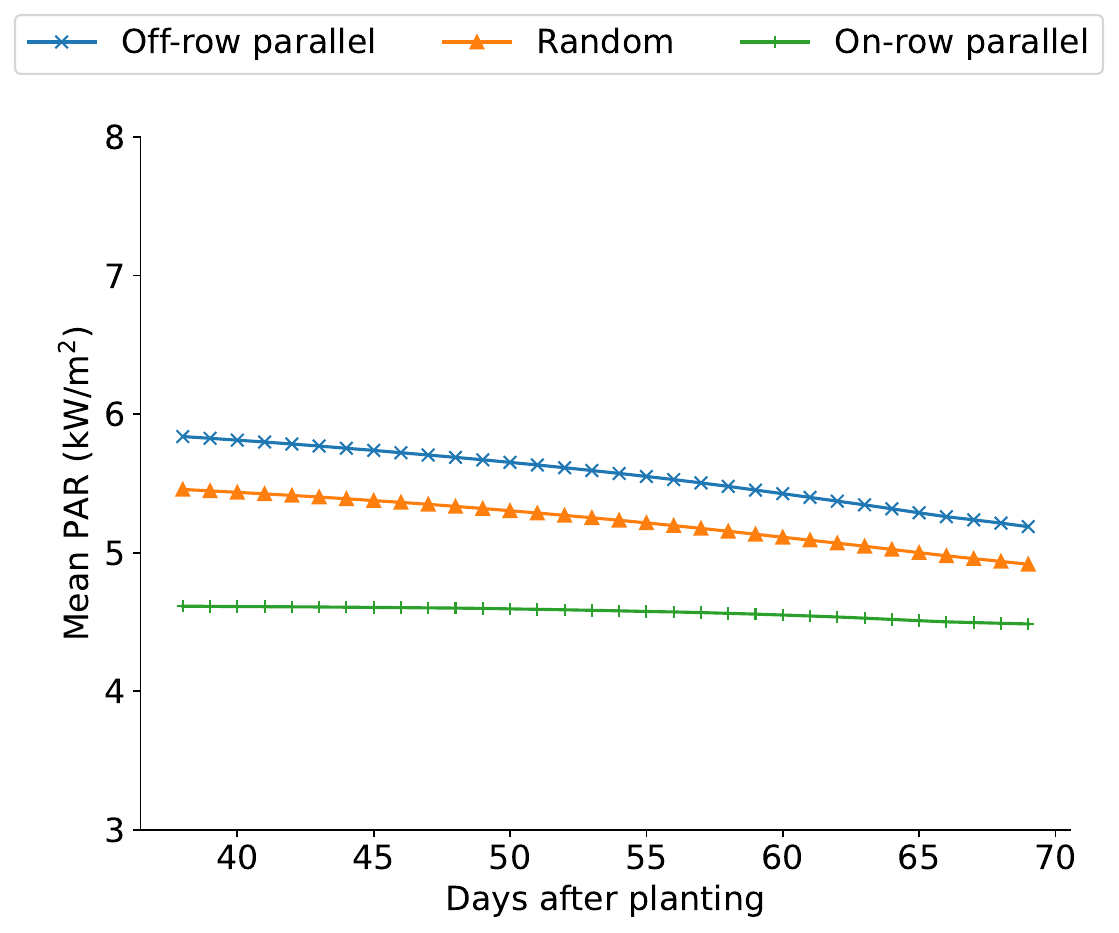}
\end{minipage}
\caption{Impact of different leaf azimuth orientations on PAR interception for a typical growing season in Ames, IA. Comparison of PAR interception among three orientations: off-row parallel, on-row parallel, and random.}
\label{fig:phyllotaxies}
\end{figure}

\subsection{Impact of Canopy Reorientation on Light Interception}

As previously discussed, some maize genotypes can adapt their architecture in response to increased planting density or altered plant distribution patterns~\citep{zhou2024genetic}. One key adaptation is canopy reorientation, where plants adjust the azimuthal orientation of their leaves to minimize shading from neighboring plants. To analyze the effect of different canopy orientations on PAR interception, we evaluated three primary arrangements:  
\begin{itemize}[topsep=2pt,itemsep=-2pt,left=6pt]
   \item \textbf{Off-row parallel:} Leaves orient perpendicular to the row direction, enhancing light capture by reducing inter-plant shading.  
    \item \textbf{On-row parallel:} Leaves align along the row direction, leading to greater overlap and reduced light capture due to shading.  
    \item \textbf{Random:} Leaves are oriented in varied directions, creating an intermediate shading effect.  
\end{itemize}

In this analysis, only the canopy orientation was varied, while all other parameters—such as planting density, row direction, and environmental conditions—were kept constant. This approach allowed us to isolate the effect of canopy reorientation on light interception. For each configuration, we analyzed PAR interception across two axes. First, we examined daily dynamics, which track how PAR interception fluctuates throughout the day as sunlight angles change. Second, we evaluated seasonal dynamics, which measure cumulative PAR interception over the growing season to capture long-term effects.

The results, shown in \figref{fig:phyllotaxies} , clearly demonstrate the impact of canopy orientation on light interception. On average, the off-row parallel orientation consistently achieved the highest PAR interception across both daily and seasonal dynamics. Our observations agree with previous simulation studies~\citep{drouet1997spatial,drouet2008does}. This configuration minimized shading from neighboring plants by positioning the leaves perpendicular to the row direction, allowing for more sunlight to reach the plant surfaces. The random orientation provided moderate light interception, as the varied leaf positions helped to reduce some shading but did not fully eliminate it. Conversely, the on-row parallel orientation resulted in the lowest PAR interception, primarily due to increased shading caused by leaf overlap when the leaves aligned along the row direction. These findings emphasize that altering canopy orientation plays a significant role in improving light interception, with the off-row parallel orientation offering the most efficient configuration for optimizing maize field productivity.
 
\begin{figure}[t!]
    \centering
    \includegraphics[width=0.95\linewidth, trim={0mm 100mm 0mm 100mm},clip]{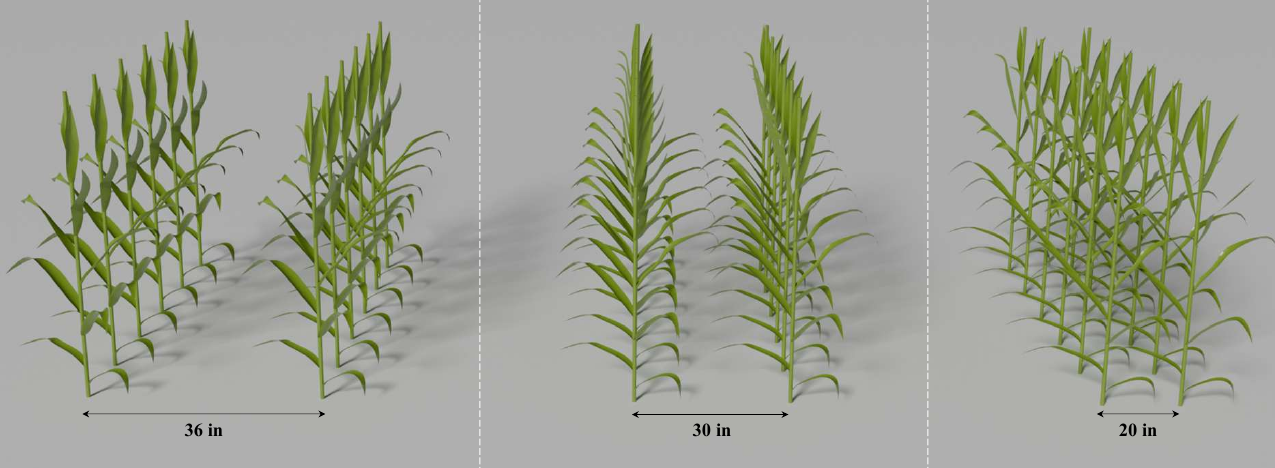}
    \includegraphics[width=1.0\linewidth, height=0.25\textheight, trim={0mm 0mm 0mm 0mm},clip]{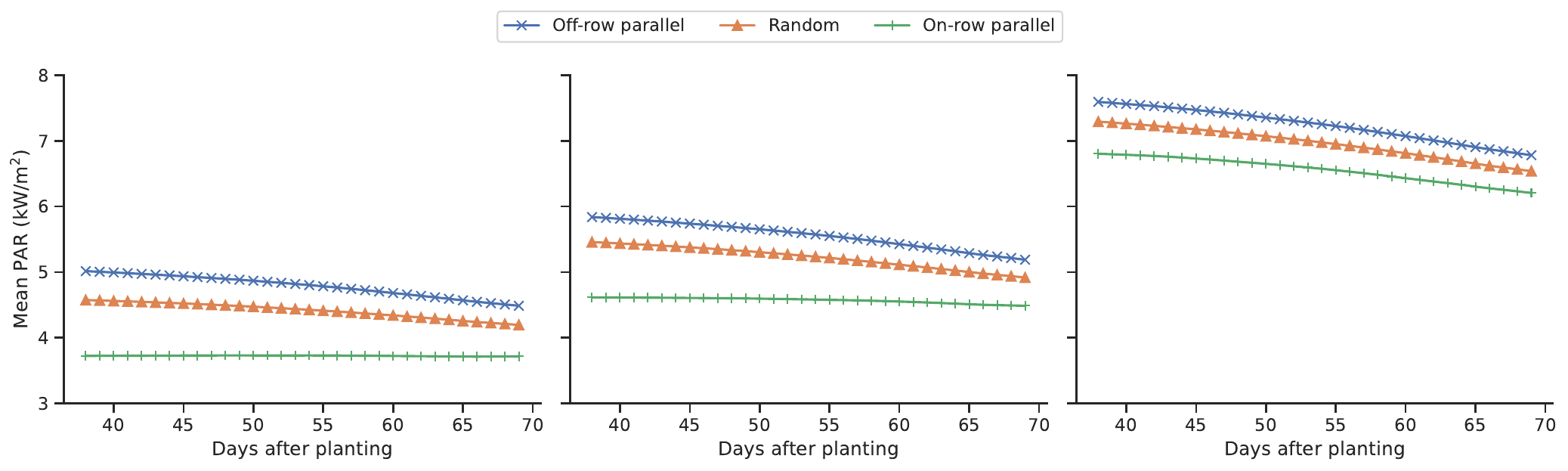}
    \caption{Effect of varying row spacing (36 inches, 30 inches, and 20 inches) on PAR interception across different leaf azimuth orientations -- off-row parallel, on-row parallel, and random over a typical growing season in maize canopies. The three plots on the bottom row represent PAR interception over the growing season for 36, 30, and 20 inch row spacing, respectively.}
    \label{fig:row spacing}
\end{figure}
\begin{figure}[t!]
    \centering
    \includegraphics[width=0.8\linewidth, trim={0mm 90mm 0mm 90mm},clip]{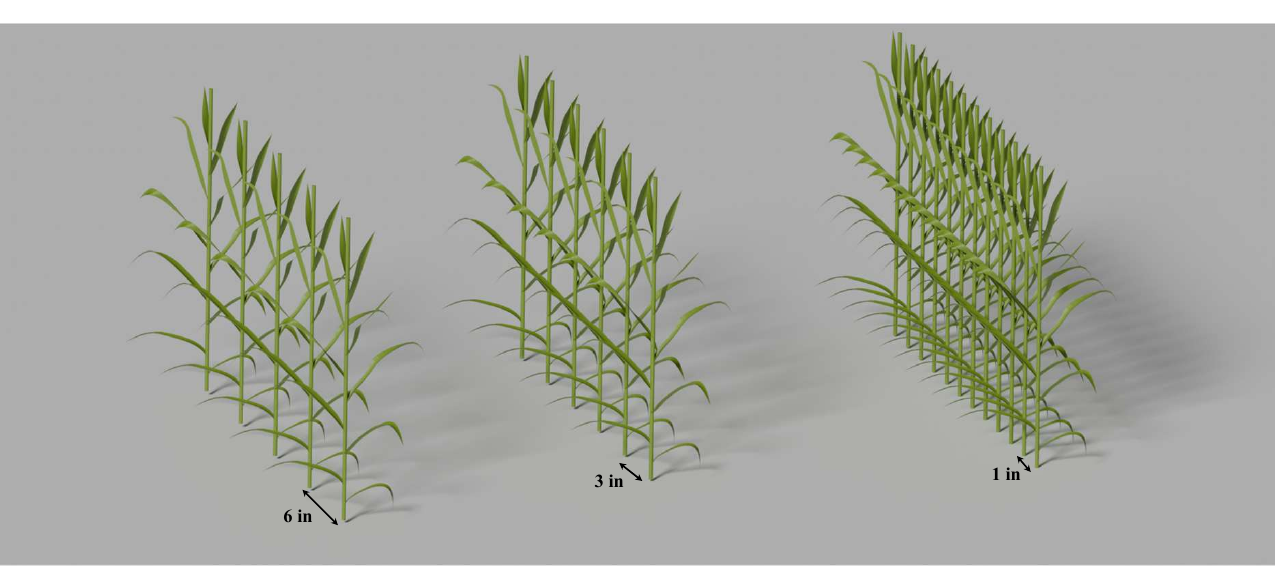}
    \includegraphics[width=1.0\linewidth, height=0.28\textheight, trim={0mm 0mm 0mm 0mm},clip]{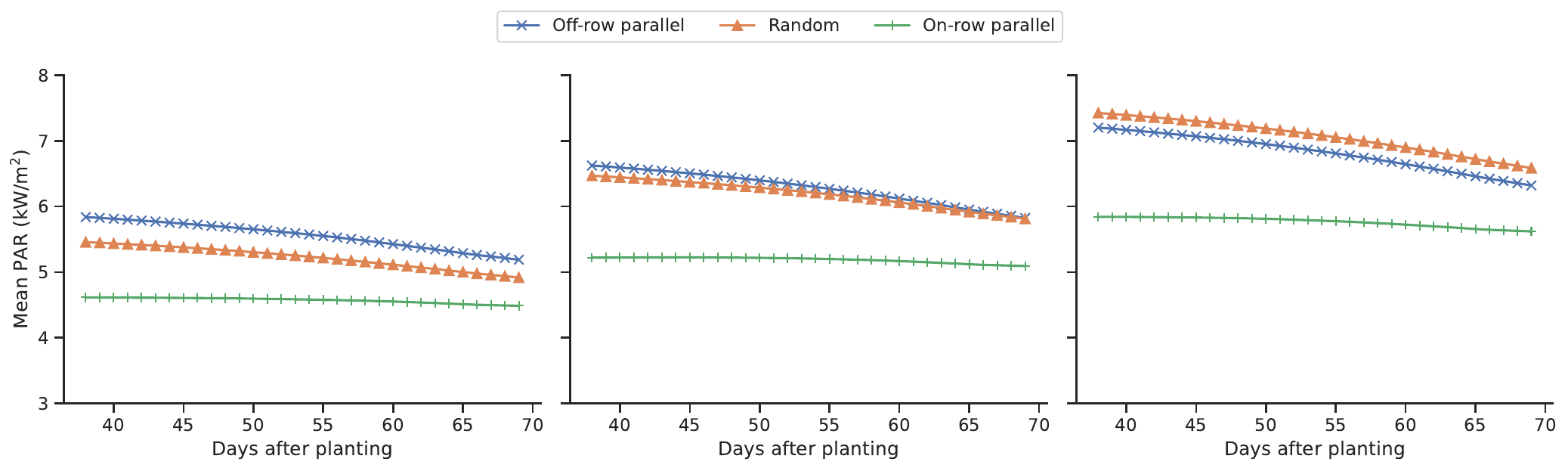}
    \caption{Impact of varying plant spacing (6 inches, 3 inches, and 1 inch) on PAR interception for different leaf azimuth orientations (on-row parallel, off-row parallel, and random) over a typical growing season in maize canopies. The three plots on the bottom row represent PAR interception over the growing season for 6, 3, and 1 inch plant spacing, respectively.}
    \label{fig:plant spacing}
\end{figure}

\subsection{Impact of Planting Density on Light Interception}

To analyze the effects of planting density on light interception, we systematically adjusted the inter-plant and inter-row distances within the maize canopy. This analysis focused on understanding how modifications to planting arrangements influenced PAR interception across different leaf orientations—off-row parallel, on-row parallel, and random—over the entire growing season.

\begin{itemize}
    \item \textbf{Varying Row Spacing:}  
    We varied the row spacing from 36 inches to 30 inches and further to 20 inches (\figref{fig:row spacing}). As expected, reducing row spacing had a significant impact on PAR interception. The off-row parallel orientation consistently achieved the highest PAR interception across all row spacings. This is attributed to its ability to minimize shading from adjacent plants by directing leaves perpendicular to the rows.  

    However, as row spacing decreased, the difference in PAR interception between the on-row parallel and the other two orientations lessened. This occurred because the off-row parallel and random orientations experienced more shading from adjacent rows, while the on-row parallel orientation was less affected by shading from nearby rows.

    \item \textbf{Varying Plant Spacing:}  
    We reduced intra-row plant spacing from 6 inches to 3 inches and finally to 1 inch (\figref{fig:plant spacing}). Unlike row spacing, reducing plant spacing had a relatively smaller impact on PAR interception per unit area. The off-row parallel orientation remained the most efficient, while the random orientation also performed better than the on-row parallel orientation, which suffered the most from increased shading caused by closer plant proximity.  

    Interestingly, as plant spacing decreased, the difference in PAR interception between the on-row parallel orientation and the other two orientations became more pronounced. This occurred because the on-row parallel orientation experienced greater shading from adjacent plants when plant spacing was reduced, while the off-row parallel and random orientations were less affected by this change.

\end{itemize}
 
These results underscore the importance of leaf azimuth orientation (canopy orientation) in optimizing light interception, particularly in high-density planting systems. The consistent superiority of the off-row parallel orientation across varying planting densities highlights its potential as a key factor in enhancing maize productivity. These trends remain consistent for simulations performed for Bisarck, North Dakota, and Thomas Co, Kansas, and are shown in \figsref{fig:bismarck}{fig:thomasco}.

\subsection{Impact of planting row directions on light interception}
\begin{figure}[t!]
    \centering
    \begin{minipage}{0.49\linewidth}
        \centering
        \includegraphics[width=0.9\linewidth, trim={40mm 80mm 40mm 82mm},clip]{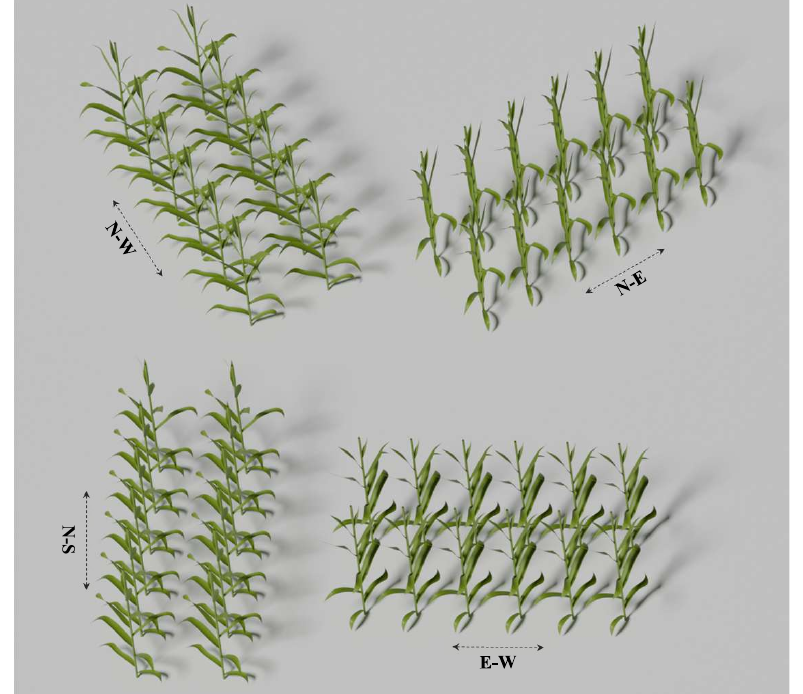}
    \end{minipage}
    \hfill
    \begin{minipage}{0.5\linewidth}
        \centering
        \includegraphics[width=\linewidth, height=0.3\textheight, trim={0mm 0mm 0mm 0mm},clip]{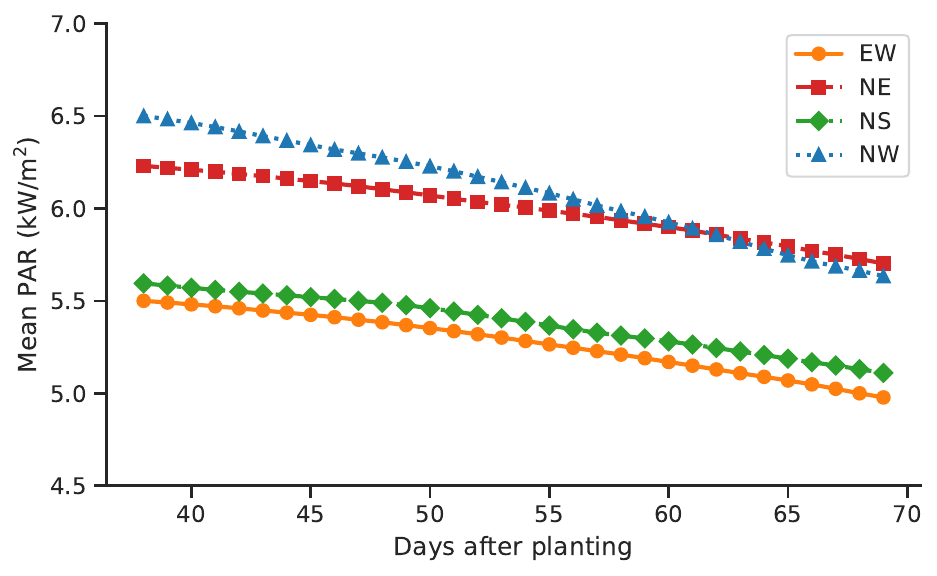}
    \end{minipage}
    \caption{Impact of different planting row orientations (east-west, north-south, northeast-southwest, and northwest-southeast) on PAR interception for maize canopies in Ames, IA, using the off-row parallel leaf orientation. The northeast-southwest (NE-SW) orientation showed the highest PAR interception, highlighting the influence of geographical factors on light capture efficiency.}
    \label{fig:row_dir}
\end{figure}

After consistently observing the superior performance of the off-row parallel orientation for light interception, we selected this configuration for a more detailed analysis of planting row directions. We explored four primary orientations: east-west (E-W), north-south (N-S), northeast-southwest (NE-SW), and northwest-southeast (NW-SE), to assess their impact on PAR interception. Our findings revealed that for Ames, IA, the NE-SW and NW-SE orientations consistently captured more PAR than the other two orientations. This variation in PAR interception is influenced by both geographical location and canopy orientation, underscoring the importance of considering regional factors when optimizing planting strategies. The results of this analysis are shown in \figref{fig:row_dir}

\section{Discussion and Conclusions}\label{sec:conclusion}

In this study, we developed and validated a framework to analyze the impact of different planting densities and leaf azimuth orientations on PAR interception in maize canopies. Our research integrated accurate 3D reconstructions of maize plants with a radiation transfer model to simulate and evaluate light interception dynamics under different planting densities, canopy orientations, and planting row directions. Our findings provide notable insights into the relationship between these factors and their effects on light use efficiency in maize, with implications for optimizing agricultural practices, recognizing that water and nutrient supply can constraint optimal plant densities and that field conditions can constrain planting row directions.

\subsection{Key Findings:}

\begin{itemize}
    \item \textbf{Leaf Azimuth Orientation:} One of the most striking findings of our study was the consistent superiority of the off-row parallel leaf azimuthal orientation in capturing PAR across all planting densities. This configuration, where the maize leaves are aligned perpendicular to the row direction, minimized shading from adjacent plants and maximized light interception. In comparison, the on-row parallel and random orientations resulted in lower PAR interception, with the on-row parallel orientation being particularly inefficient due to significant shading from neighboring plants within the same row. These results highlight the importance of leaf orientation in enhancing light capture, especially in high-density planting systems.

    \item \textbf{Planting Density:} Our analysis revealed that row spacing had a more significant effect on PAR interception than plant spacing. Reducing row spacing from 36 inches to 20 inches resulted in a substantial increase in light interception. On the other hand, reducing intra-row plant spacing had a relatively smaller impact on PAR interception, especially in the off-row parallel configuration. These results suggest that optimizing row spacing is more critical for improving light interception than adjusting plant spacing. That said, as discussed above, other factors, including the availability of appropriate planting equipment also constrain row spacing. The robustness of the off-row parallel orientation, even under high-density conditions, further supports its potential for maximizing light capture in dense planting scenarios.

    \item \textbf{Row Direction:} In terms of row orientation, our study found that the northeast-southwest (NE-SW) and northwest-southeast (NW-SE) orientations consistently captured more PAR compared to the east-west (E-W) and north-south (N-S) orientations for Ames, IA. This variation in PAR interception is influenced by the leaf azimuthal orientation and geographical factors, which are important considerations when designing planting strategies to maximize light capture.
\end{itemize}

The insights gained from this study have useful implications for maize cultivation practices. By carefully optimizing planting density, leaf orientation, and row orientation, farmers and breeders can improve light interception, which is directly linked to increased photosynthetic efficiency and higher crop yields. The off-row parallel orientation, in particular, offers a promising strategy for maximizing light capture in both conventional and high-density planting systems. Furthermore, the findings suggest that breeding programs aimed at adjusting genetic traits for azimuthal leaf orientations could enhance light interception, particularly in conditions where maximizing sunlight exposure is critical. This approach could easily be applied to optimize other canopy architecture, such as vertical leaf angle, leaf, number, vertical spacing, length, and width. 

Future research could focus on exploring the genetic basis of leaf reorientation and its potential for breeding programs aimed at improving light interception and yield in dense planting systems. Additionally, extending the framework to other crops and environmental conditions would further validate its applicability and provide broader insights into optimizing agricultural productivity. Overall, our integrated framework provides a robust tool for analyzing and optimizing light interception in maize canopies, contributing to the development of more efficient and productive agricultural systems. 

\section*{Acknowledgements}
This work was supported by the AI Institute for Resilient Agriculture (USDA-NIFA 2021-67021-35329), a grant from the National Science Foundation (No. IOS-1238142) to PSS and others, and Iowa State University's Plant Science Institute.  YL was supported in part by NSF grants (DBI-1661475 and IOS-1842097) to PSS and others and a scholarship from the China Scholarship Council.

\section*{Author Contributions}
Credit: Conceptualization: NS, TZJ, PSS, AK, BG; Data curation: TZJ, YZ, YL; Formal Analysis: NS; Funding acquisition: PSS, AK, BG; Investigation: NS, TZJ, YZ, YL; Methodology: NS, TZJ, AK, BG; Project administration: PS, AK, BG; Resources: PS, AK, BG; Software: NS; Supervision: PSS, AK, BG; Validation: NS, YZ, YL; Visualization: NS; Writing – original draft: NS; Writing – review \& editing: NS, TZJ, AB, PSS, AK, BG

\section*{Conflict of Interest}
The authors declare that there are no competing interests regarding the publication of this article.

\section*{Data Availability}
The data used for analysis in this manuscript is available upon request.

\bibliographystyle{elsarticle-num-names} 
\bibliography{references}

\begin{thebibliography}{46}
\expandafter\ifx\csname natexlab\endcsname\relax\def\natexlab#1{#1}\fi
\providecommand{\url}[1]{\texttt{#1}}
\providecommand{\href}[2]{#2}
\providecommand{\path}[1]{#1}
\providecommand{\DOIprefix}{doi:}
\providecommand{\ArXivprefix}{arXiv:}
\providecommand{\URLprefix}{URL: }
\providecommand{\Pubmedprefix}{pmid:}
\providecommand{\doi}[1]{\href{http://dx.doi.org/#1}{\path{#1}}}
\providecommand{\Pubmed}[1]{\href{pmid:#1}{\path{#1}}}
\providecommand{\bibinfo}[2]{#2}
\ifx\xfnm\relax \def\xfnm[#1]{\unskip,\space#1}\fi
\bibitem[{Erenstein et~al.(2022)Erenstein, Jaleta, Sonder, Mottaleb, and Prasanna}]{erenstein2022global}
\bibinfo{author}{O.~Erenstein}, \bibinfo{author}{M.~Jaleta}, \bibinfo{author}{K.~Sonder}, \bibinfo{author}{K.~Mottaleb}, \bibinfo{author}{B.~M. Prasanna},
\newblock \bibinfo{title}{Global maize production, consumption and trade: trends and r\&d implications},
\newblock \bibinfo{journal}{Food security} \bibinfo{volume}{14} (\bibinfo{year}{2022}) \bibinfo{pages}{1295--1319}.
\bibitem[{Canton(2021)}]{canton2021food}
\bibinfo{author}{H.~Canton},
\newblock \bibinfo{title}{Food and agriculture organization ({FAO}) of the {United} {Nations}},
\newblock in: \bibinfo{booktitle}{The Europa directory of international organizations}, \bibinfo{publisher}{Routledge}, \bibinfo{year}{2021}, pp. \bibinfo{pages}{297--305}.
\bibitem[{Tokatlidis and Koutroubas(2004)}]{tokatlidis2004review}
\bibinfo{author}{I.~Tokatlidis}, \bibinfo{author}{S.~Koutroubas},
\newblock \bibinfo{title}{A review of maize hybrids’ dependence on high plant populations and its implications for crop yield stability},
\newblock \bibinfo{journal}{Field Crops Research} \bibinfo{volume}{88} (\bibinfo{year}{2004}) \bibinfo{pages}{103--114}.
\bibitem[{Duvick et~al.(2004)Duvick, Smith, Cooper et~al.}]{duvick2004long}
\bibinfo{author}{D.~N. Duvick}, \bibinfo{author}{J.~Smith}, \bibinfo{author}{M.~Cooper}, et~al.,
\newblock \bibinfo{title}{Long-term selection in a commercial hybrid maize breeding program},
\newblock \bibinfo{journal}{Plant breeding reviews} \bibinfo{volume}{24} (\bibinfo{year}{2004}) \bibinfo{pages}{109--152}.
\bibitem[{Berzsenyi and Tokatlidis(2012)}]{berzsenyi2012density}
\bibinfo{author}{Z.~Berzsenyi}, \bibinfo{author}{I.~Tokatlidis},
\newblock \bibinfo{title}{Density dependence rather than maturity determines hybrid selection in dryland maize production},
\newblock \bibinfo{journal}{Agronomy journal} \bibinfo{volume}{104} (\bibinfo{year}{2012}) \bibinfo{pages}{331--336}.
\bibitem[{Gonzalez et~al.(2018)Gonzalez, Tollenaar, Bowman, Good, and Lee}]{gonzalez2018maize}
\bibinfo{author}{V.~Gonzalez}, \bibinfo{author}{M.~Tollenaar}, \bibinfo{author}{A.~Bowman}, \bibinfo{author}{B.~Good}, \bibinfo{author}{E.~Lee},
\newblock \bibinfo{title}{Maize yield potential and density tolerance},
\newblock \bibinfo{journal}{Crop Science} \bibinfo{volume}{58} (\bibinfo{year}{2018}) \bibinfo{pages}{472--485}.
\bibitem[{Assefa et~al.(2018)Assefa, Carter, Hinds, Bhalla, Schon, Jeschke, Paszkiewicz, Smith, and Ciampitti}]{assefa2018analysis}
\bibinfo{author}{Y.~Assefa}, \bibinfo{author}{P.~Carter}, \bibinfo{author}{M.~Hinds}, \bibinfo{author}{G.~Bhalla}, \bibinfo{author}{R.~Schon}, \bibinfo{author}{M.~Jeschke}, \bibinfo{author}{S.~Paszkiewicz}, \bibinfo{author}{S.~Smith}, \bibinfo{author}{I.~A. Ciampitti},
\newblock \bibinfo{title}{Analysis of long term study indicates both agronomic optimal plant density and increase maize yield per plant contributed to yield gain},
\newblock \bibinfo{journal}{Scientific reports} \bibinfo{volume}{8} (\bibinfo{year}{2018}) \bibinfo{pages}{4937}.
\bibitem[{Duvick and Cassman(1999)}]{duvick1999post}
\bibinfo{author}{D.~N. Duvick}, \bibinfo{author}{K.~G. Cassman},
\newblock \bibinfo{title}{Post--green revolution trends in yield potential of temperate maize in the north-central {United States}},
\newblock \bibinfo{journal}{Crop science} \bibinfo{volume}{39} (\bibinfo{year}{1999}) \bibinfo{pages}{1622--1630}.
\bibitem[{Duvick et~al.(2005)}]{duvick2005genetic}
\bibinfo{author}{D.~Duvick}, et~al.,
\newblock \bibinfo{title}{Genetic progress in yield of {United States} maize ({Zea mays L.})},
\newblock \bibinfo{journal}{Maydica} \bibinfo{volume}{50} (\bibinfo{year}{2005}) \bibinfo{pages}{193}.
\bibitem[{Earley et~al.(1966)Earley, Miller, Reichert, Hageman, and Seif}]{earley1966effect}
\bibinfo{author}{E.~Earley}, \bibinfo{author}{R.~Miller}, \bibinfo{author}{G.~Reichert}, \bibinfo{author}{R.~Hageman}, \bibinfo{author}{R.~Seif},
\newblock \bibinfo{title}{Effect of shade on maize production under field conditions},
\newblock \bibinfo{journal}{Crop Science} \bibinfo{volume}{6} (\bibinfo{year}{1966}) \bibinfo{pages}{1--7}.
\bibitem[{Reed et~al.(1988)Reed, Singletary, Schussler, Williamson, and Christy}]{reed1988shading}
\bibinfo{author}{A.~Reed}, \bibinfo{author}{G.~Singletary}, \bibinfo{author}{J.~Schussler}, \bibinfo{author}{D.~Williamson}, \bibinfo{author}{A.~Christy},
\newblock \bibinfo{title}{Shading effects on dry matter and nitrogen partitioning, kernel number, and yield of maize},
\newblock \bibinfo{journal}{Crop Science} \bibinfo{volume}{28} (\bibinfo{year}{1988}) \bibinfo{pages}{819--825}.
\bibitem[{Lambert and Johnson(1978)}]{lambert1978leaf}
\bibinfo{author}{R.~Lambert}, \bibinfo{author}{R.~Johnson},
\newblock \bibinfo{title}{Leaf angle, tassel morphology, and the performance of maize hybrids},
\newblock \bibinfo{journal}{Crop Science} \bibinfo{volume}{18} (\bibinfo{year}{1978}) \bibinfo{pages}{499--502}.
\bibitem[{Toler et~al.(1999)Toler, Murdock, Stapleton, and Wallace}]{toler1999corn}
\bibinfo{author}{J.~Toler}, \bibinfo{author}{E.~Murdock}, \bibinfo{author}{G.~Stapleton}, \bibinfo{author}{S.~Wallace},
\newblock \bibinfo{title}{Corn leaf orientation effects on light interception, intraspecific competition, and grain yields},
\newblock \bibinfo{journal}{Journal of production agriculture} \bibinfo{volume}{12} (\bibinfo{year}{1999}) \bibinfo{pages}{396--399}.
\bibitem[{Hammer et~al.(2009)Hammer, Dong, McLean, Doherty, Messina, Schussler, Zinselmeier, Paszkiewicz, and Cooper}]{hammer2009can}
\bibinfo{author}{G.~L. Hammer}, \bibinfo{author}{Z.~Dong}, \bibinfo{author}{G.~McLean}, \bibinfo{author}{A.~Doherty}, \bibinfo{author}{C.~Messina}, \bibinfo{author}{J.~Schussler}, \bibinfo{author}{C.~Zinselmeier}, \bibinfo{author}{S.~Paszkiewicz}, \bibinfo{author}{M.~Cooper},
\newblock \bibinfo{title}{Can changes in canopy and/or root system architecture explain historical maize yield trends in the us corn belt?},
\newblock \bibinfo{journal}{Crop Science} \bibinfo{volume}{49} (\bibinfo{year}{2009}) \bibinfo{pages}{299--312}.
\bibitem[{Al-Naggar et~al.(2015)Al-Naggar, Shabana, Atta, and Al-Khalil}]{al2015maize}
\bibinfo{author}{A.~M.~M. Al-Naggar}, \bibinfo{author}{R.~A. Shabana}, \bibinfo{author}{M.~M. Atta}, \bibinfo{author}{T.~H. Al-Khalil},
\newblock \bibinfo{title}{Maize response to elevated plant density combined with lowered n-fertilizer rate is genotype-dependent},
\newblock \bibinfo{journal}{The Crop Journal} \bibinfo{volume}{3} (\bibinfo{year}{2015}) \bibinfo{pages}{96--109}.
\bibitem[{Sangoi(2001)}]{sangoi2001understanding}
\bibinfo{author}{L.~Sangoi},
\newblock \bibinfo{title}{Understanding plant density effects on maize growth and development: {A}n important issue to maximize grain yield},
\newblock \bibinfo{journal}{Ci{\^e}ncia rural} \bibinfo{volume}{31} (\bibinfo{year}{2001}) \bibinfo{pages}{159--168}.
\bibitem[{Wimalasekera(2019)}]{wimalasekera2019effect}
\bibinfo{author}{R.~Wimalasekera},
\newblock \bibinfo{title}{Effect of light intensity on photosynthesis},
\newblock \bibinfo{journal}{Photosynthesis, productivity and environmental stress}  (\bibinfo{year}{2019}) \bibinfo{pages}{65--73}.
\bibitem[{Niinemets(2010)}]{niinemets2010review}
\bibinfo{author}{{\"U}.~Niinemets},
\newblock \bibinfo{title}{A review of light interception in plant stands from leaf to canopy in different plant functional types and in species with varying shade tolerance},
\newblock \bibinfo{journal}{Ecological Research} \bibinfo{volume}{25} (\bibinfo{year}{2010}) \bibinfo{pages}{693--714}.
\bibitem[{Zhu et~al.(2012)Zhu, Song, and Ort}]{zhu2012elements}
\bibinfo{author}{X.-G. Zhu}, \bibinfo{author}{Q.~Song}, \bibinfo{author}{D.~R. Ort},
\newblock \bibinfo{title}{Elements of a dynamic systems model of canopy photosynthesis},
\newblock \bibinfo{journal}{Current opinion in plant biology} \bibinfo{volume}{15} (\bibinfo{year}{2012}) \bibinfo{pages}{237--244}.
\bibitem[{Hammer et~al.(2002)Hammer, Kropff, Sinclair, and Porter}]{hammer2002future}
\bibinfo{author}{G.~Hammer}, \bibinfo{author}{M.~Kropff}, \bibinfo{author}{T.~Sinclair}, \bibinfo{author}{J.~Porter},
\newblock \bibinfo{title}{Future contributions of crop modelling—from heuristics and supporting decision making to understanding genetic regulation and aiding crop improvement},
\newblock \bibinfo{journal}{European Journal of Agronomy} \bibinfo{volume}{18} (\bibinfo{year}{2002}) \bibinfo{pages}{15--31}.
\bibitem[{R{\"o}tter et~al.(2015)R{\"o}tter, Tao, H{\"o}hn, and Palosuo}]{rotter2015use}
\bibinfo{author}{R.~P. R{\"o}tter}, \bibinfo{author}{F.~Tao}, \bibinfo{author}{J.~G. H{\"o}hn}, \bibinfo{author}{T.~Palosuo},
\newblock \bibinfo{title}{Use of crop simulation modelling to aid ideotype design of future cereal cultivars},
\newblock \bibinfo{journal}{Journal of experimental botany} \bibinfo{volume}{66} (\bibinfo{year}{2015}) \bibinfo{pages}{3463--3476}.
\bibitem[{Maddonni et~al.(2002)Maddonni, Otegui, Andrieu, Chelle, and Casal}]{maddonni2002maize}
\bibinfo{author}{G.~A. Maddonni}, \bibinfo{author}{M.~E. Otegui}, \bibinfo{author}{B.~Andrieu}, \bibinfo{author}{M.~Chelle}, \bibinfo{author}{J.~J. Casal},
\newblock \bibinfo{title}{Maize leaves turn away from neighbors},
\newblock \bibinfo{journal}{Plant physiology} \bibinfo{volume}{130} (\bibinfo{year}{2002}) \bibinfo{pages}{1181--1189}.
\bibitem[{Drouet and Moulia(1997)}]{drouet1997spatial}
\bibinfo{author}{J.-L. Drouet}, \bibinfo{author}{B.~Moulia},
\newblock \bibinfo{title}{Spatial re-orientation of maize leaves affected by initial plant orientation and density},
\newblock \bibinfo{journal}{Agricultural and Forest meteorology} \bibinfo{volume}{88} (\bibinfo{year}{1997}) \bibinfo{pages}{85--100}.
\bibitem[{Drouet and Kiniry(2008)}]{drouet2008does}
\bibinfo{author}{J.-L. Drouet}, \bibinfo{author}{J.~Kiniry},
\newblock \bibinfo{title}{Does spatial arrangement of {3D} plants affect light transmission and extinction coefficient within maize crops?},
\newblock \bibinfo{journal}{Field Crops Research} \bibinfo{volume}{107} (\bibinfo{year}{2008}) \bibinfo{pages}{62--69}.
\bibitem[{Maddonni et~al.(2001)Maddonni, Chelle, Drouet, and Andrieu}]{maddonni2001light}
\bibinfo{author}{G.~Maddonni}, \bibinfo{author}{M.~Chelle}, \bibinfo{author}{J.-L. Drouet}, \bibinfo{author}{B.~Andrieu},
\newblock \bibinfo{title}{Light interception of contrasting azimuth canopies under square and rectangular plant spatial distributions: simulations and crop measurements},
\newblock \bibinfo{journal}{Field Crops Research} \bibinfo{volume}{70} (\bibinfo{year}{2001}) \bibinfo{pages}{1--13}.
\bibitem[{Zhou et~al.(2024)Zhou, Kusmec, and Schnable}]{zhou2024genetic}
\bibinfo{author}{Y.~Zhou}, \bibinfo{author}{A.~Kusmec}, \bibinfo{author}{P.~S. Schnable},
\newblock \bibinfo{title}{Genetic regulation of self-organizing azimuthal canopy orientations and their impacts on light interception in maize},
\newblock \bibinfo{journal}{The Plant Cell} \bibinfo{volume}{36} (\bibinfo{year}{2024}) \bibinfo{pages}{1600--1621}.
\bibitem[{Drouet and Pag{\`e}s(2003)}]{drouet2003graal}
\bibinfo{author}{J.-L. Drouet}, \bibinfo{author}{L.~Pag{\`e}s},
\newblock \bibinfo{title}{Graal: {A} model of growth, architecture and carbon allocation during the vegetative phase of the whole maize plant: model description and parameterisation},
\newblock \bibinfo{journal}{Ecological Modelling} \bibinfo{volume}{165} (\bibinfo{year}{2003}) \bibinfo{pages}{147--173}.
\bibitem[{Song et~al.(2008)Song, Birch, and Hanan}]{song2008analysis}
\bibinfo{author}{Y.~Song}, \bibinfo{author}{C.~Birch}, \bibinfo{author}{J.~Hanan},
\newblock \bibinfo{title}{Analysis of maize canopy development under water stress and incorporation into the adel-maize model},
\newblock \bibinfo{journal}{Functional Plant Biology} \bibinfo{volume}{35} (\bibinfo{year}{2008}) \bibinfo{pages}{925--935}.
\bibitem[{Kim et~al.(2020)Kim, Kang, Hwang, Kim, Kim, Park, and Son}]{kim2020use}
\bibinfo{author}{D.~Kim}, \bibinfo{author}{W.~H. Kang}, \bibinfo{author}{I.~Hwang}, \bibinfo{author}{J.~Kim}, \bibinfo{author}{J.~H. Kim}, \bibinfo{author}{K.~S. Park}, \bibinfo{author}{J.~E. Son},
\newblock \bibinfo{title}{Use of structurally-accurate {3D} plant models for estimating light interception and photosynthesis of sweet pepper (capsicum annuum) plants},
\newblock \bibinfo{journal}{Computers and Electronics in Agriculture} \bibinfo{volume}{177} (\bibinfo{year}{2020}) \bibinfo{pages}{105689}.
\bibitem[{Paturkar et~al.(2021)Paturkar, Sen~Gupta, and Bailey}]{paturkar2021making}
\bibinfo{author}{A.~Paturkar}, \bibinfo{author}{G.~Sen~Gupta}, \bibinfo{author}{D.~Bailey},
\newblock \bibinfo{title}{Making use of {3D} models for plant physiognomic analysis: {A} review},
\newblock \bibinfo{journal}{Remote Sensing} \bibinfo{volume}{13} (\bibinfo{year}{2021}) \bibinfo{pages}{2232}.
\bibitem[{Okura(2022)}]{okura20223d}
\bibinfo{author}{F.~Okura},
\newblock \bibinfo{title}{3d modeling and reconstruction of plants and trees: {A} cross-cutting review across computer graphics, vision, and plant phenotyping},
\newblock \bibinfo{journal}{Breeding Science} \bibinfo{volume}{72} (\bibinfo{year}{2022}) \bibinfo{pages}{31--47}.
\bibitem[{Gu et~al.(2022)Gu, Wen, Xu, Lu, Yu, Guo, and Zhao}]{gu2022use}
\bibinfo{author}{S.~Gu}, \bibinfo{author}{W.~Wen}, \bibinfo{author}{T.~Xu}, \bibinfo{author}{X.~Lu}, \bibinfo{author}{Z.~Yu}, \bibinfo{author}{X.~Guo}, \bibinfo{author}{C.~Zhao},
\newblock \bibinfo{title}{Use of {3D} modeling to refine predictions of canopy light utilization: {A} comparative study on canopy photosynthesis models with different dimensions},
\newblock \bibinfo{journal}{Frontiers in Plant Science} \bibinfo{volume}{13} (\bibinfo{year}{2022}) \bibinfo{pages}{735981}.
\bibitem[{Xiao et~al.(2023)Xiao, Fei, Li, Zhang, Chen, Xu, Cai, Bi, Guo, Li et~al.}]{xiao2023importance}
\bibinfo{author}{S.~Xiao}, \bibinfo{author}{S.~Fei}, \bibinfo{author}{Q.~Li}, \bibinfo{author}{B.~Zhang}, \bibinfo{author}{H.~Chen}, \bibinfo{author}{D.~Xu}, \bibinfo{author}{Z.~Cai}, \bibinfo{author}{K.~Bi}, \bibinfo{author}{Y.~Guo}, \bibinfo{author}{B.~Li}, et~al.,
\newblock \bibinfo{title}{The importance of using realistic {3D} canopy models to calculate light interception in the field},
\newblock \bibinfo{journal}{Plant Phenomics} \bibinfo{volume}{5} (\bibinfo{year}{2023}) \bibinfo{pages}{0082}.
\bibitem[{Sultana et~al.(2023)Sultana, Dev, Xin, Han, Feng, Lei, Yang, Wang, Li, Wang et~al.}]{sultana2023competition}
\bibinfo{author}{F.~Sultana}, \bibinfo{author}{W.~Dev}, \bibinfo{author}{M.~Xin}, \bibinfo{author}{Y.~Han}, \bibinfo{author}{L.~Feng}, \bibinfo{author}{Y.~Lei}, \bibinfo{author}{B.~Yang}, \bibinfo{author}{G.~Wang}, \bibinfo{author}{X.~Li}, \bibinfo{author}{Z.~Wang}, et~al.,
\newblock \bibinfo{title}{Competition for light interception in different plant canopy characteristics of diverse cotton cultivars},
\newblock \bibinfo{journal}{Genes} \bibinfo{volume}{14} (\bibinfo{year}{2023}) \bibinfo{pages}{364}.
\bibitem[{Zhao et~al.(2024)Zhao, Qi, Yu, Yuan, and Huang}]{zhao2024fine}
\bibinfo{author}{X.~Zhao}, \bibinfo{author}{J.~Qi}, \bibinfo{author}{Z.~Yu}, \bibinfo{author}{L.~Yuan}, \bibinfo{author}{H.~Huang},
\newblock \bibinfo{title}{Fine-scale quantification of absorbed photosynthetically active radiation (apar) in plantation forests with {3D} radiative transfer modeling and lidar data},
\newblock \bibinfo{journal}{Plant Phenomics} \bibinfo{volume}{6} (\bibinfo{year}{2024}) \bibinfo{pages}{0166}.
\bibitem[{Song et~al.(2023)Song, Liu, Bu, and Zhu}]{song2023quantifying}
\bibinfo{author}{Q.~Song}, \bibinfo{author}{F.~Liu}, \bibinfo{author}{H.~Bu}, \bibinfo{author}{X.-G. Zhu},
\newblock \bibinfo{title}{Quantifying contributions of different factors to canopy photosynthesis in 2 maize varieties: development of a novel {3D} canopy modeling pipeline},
\newblock \bibinfo{journal}{Plant Phenomics} \bibinfo{volume}{5} (\bibinfo{year}{2023}) \bibinfo{pages}{0075}.
\bibitem[{Zhu et~al.(2020)Zhu, Liu, Xie, Guo, Li, and Ma}]{zhu2020quantification}
\bibinfo{author}{B.~Zhu}, \bibinfo{author}{F.~Liu}, \bibinfo{author}{Z.~Xie}, \bibinfo{author}{Y.~Guo}, \bibinfo{author}{B.~Li}, \bibinfo{author}{Y.~Ma},
\newblock \bibinfo{title}{Quantification of light interception within image-based 3-d reconstruction of sole and intercropped canopies over the entire growth season},
\newblock \bibinfo{journal}{Annals of botany} \bibinfo{volume}{126} (\bibinfo{year}{2020}) \bibinfo{pages}{701--712}.
\bibitem[{Leiboff et~al.(2015)Leiboff, Li, Hu, Todt, Yang, Li, Yu, Muehlbauer, Timmermans, Yu et~al.}]{leiboff2015genetic}
\bibinfo{author}{S.~Leiboff}, \bibinfo{author}{X.~Li}, \bibinfo{author}{H.-C. Hu}, \bibinfo{author}{N.~Todt}, \bibinfo{author}{J.~Yang}, \bibinfo{author}{X.~Li}, \bibinfo{author}{X.~Yu}, \bibinfo{author}{G.~J. Muehlbauer}, \bibinfo{author}{M.~C. Timmermans}, \bibinfo{author}{J.~Yu}, et~al.,
\newblock \bibinfo{title}{Genetic control of morphometric diversity in the maize shoot apical meristem},
\newblock \bibinfo{journal}{Nature Communications} \bibinfo{volume}{6} (\bibinfo{year}{2015}) \bibinfo{pages}{8974}.
\bibitem[{{FARO Technologies, Inc.}(2024)}]{faro_scene}
\bibinfo{author}{{FARO Technologies, Inc.}}, \bibinfo{title}{Faro scene software}, \bibinfo{year}{2024}. \URLprefix \url{https://www.faro.com/en/Products/Software/SCENE-Software}.
\bibitem[{CloudCompare(lpha)}]{cloudcompare}
\bibinfo{author}{CloudCompare}, \bibinfo{year}{(version 2.13.alpha)}. \URLprefix \url{http://www.cloudcompare.org/}.
\bibitem[{Miao et~al.(2021)Miao, Wen, Li, Wu, Zhu, and Guo}]{miao2021label3dmaize}
\bibinfo{author}{T.~Miao}, \bibinfo{author}{W.~Wen}, \bibinfo{author}{Y.~Li}, \bibinfo{author}{S.~Wu}, \bibinfo{author}{C.~Zhu}, \bibinfo{author}{X.~Guo},
\newblock \bibinfo{title}{Label3dmaize: toolkit for {3D} point cloud data annotation of maize shoots},
\newblock \bibinfo{journal}{GigaScience} \bibinfo{volume}{10} (\bibinfo{year}{2021}) \bibinfo{pages}{giab031}.
\bibitem[{Bernardini et~al.(1999)Bernardini, Mittleman, Rushmeier, Silva, and Taubin}]{bernardini1999ball}
\bibinfo{author}{F.~Bernardini}, \bibinfo{author}{J.~Mittleman}, \bibinfo{author}{H.~Rushmeier}, \bibinfo{author}{C.~Silva}, \bibinfo{author}{G.~Taubin},
\newblock \bibinfo{title}{The ball-pivoting algorithm for surface reconstruction},
\newblock \bibinfo{journal}{IEEE transactions on visualization and computer graphics} \bibinfo{volume}{5} (\bibinfo{year}{1999}) \bibinfo{pages}{349--359}.
\bibitem[{Cignoni et~al.(2008)Cignoni, Callieri, Corsini, Dellepiane, Ganovelli, and Ranzuglia}]{LocalChapterEvents:ItalChap:ItalianChapConf2008:129-136}
\bibinfo{author}{P.~Cignoni}, \bibinfo{author}{M.~Callieri}, \bibinfo{author}{M.~Corsini}, \bibinfo{author}{M.~Dellepiane}, \bibinfo{author}{F.~Ganovelli}, \bibinfo{author}{G.~Ranzuglia},
\newblock \bibinfo{title}{{MeshLab:} {An} open-source mesh processing tool},
\newblock in: \bibinfo{editor}{V.~Scarano}, \bibinfo{editor}{R.~D. Chiara}, \bibinfo{editor}{U.~Erra} (Eds.), \bibinfo{booktitle}{Eurographics Italian Chapter Conference}, \bibinfo{publisher}{The Eurographics Association}, \bibinfo{year}{2008}, pp. \bibinfo{pages}{1--10}. \DOIprefix\doi{10.2312/LocalChapterEvents/ItalChap/ItalianChapConf2008/129-136}.
\bibitem[{Bailey(2019)}]{bailey2019helios}
\bibinfo{author}{B.~N. Bailey},
\newblock \bibinfo{title}{{Helios:} {A} scalable {3D} plant and environmental biophysical modeling framework},
\newblock \bibinfo{journal}{Frontiers in Plant Science} \bibinfo{volume}{10} (\bibinfo{year}{2019}) \bibinfo{pages}{1185}.
\bibitem[{Bailey(2018)}]{bailey2018reverse}
\bibinfo{author}{B.~N. Bailey},
\newblock \bibinfo{title}{A reverse ray-tracing method for modelling the net radiative flux in leaf-resolving plant canopy simulations},
\newblock \bibinfo{journal}{Ecological Modelling} \bibinfo{volume}{368} (\bibinfo{year}{2018}) \bibinfo{pages}{233--245}.
\bibitem[{Earl and Tollenaar(1997)}]{earl1997maize}
\bibinfo{author}{H.~Earl}, \bibinfo{author}{M.~Tollenaar},
\newblock \bibinfo{title}{Maize leaf absorptance of photosynthetically active radiation and its estimation using a chlorophyll meter},
\newblock \bibinfo{journal}{Crop Science} \bibinfo{volume}{37} (\bibinfo{year}{1997}) \bibinfo{pages}{436--440}.

\end{thebibliography}

\clearpage
\appendix
\renewcommand{\thefigure}{A.\arabic{figure}}
\setcounter{figure}{0}
\setcounter{table}{0}

\section{Input Parameters for the Helios 3D Modeling Framework}

\begin{table}[h!]
    \centering
    \caption{Input parameters for the virtual framework.}
    \begin{tabular}{ll}
        \toprule
        \textbf{Parameter} & \textbf{Description} \\
        \midrule
        3D Canopy Models & ply format representation of maize plants \\
        Radiative Transfer Model & Reverse ray tracing method for PAR (400-700 nm) \\
        Leaf Reflectance & Uniform 0.1 across all bands \\
        Leaf Transmittance & Uniform 0.1 across all bands \\
        Simulation Time & Daily from 07:00 to 20:00, July 15 - August 15 \\
        Scattering Iterations & 5 iterations for light interactions \\
        Boundary Conditions & Periodic to minimize edge effects \\
        \bottomrule
    \end{tabular}
    \label{table:supplementary_input_parameters}
\end{table}
\clearpage
\section{Varying planting densities for other locations}

\begin{figure}[h!]
    \centering
    \includegraphics[width=\linewidth, height=0.65\textheight, trim={0mm 105mm 5mm 0mm},clip]{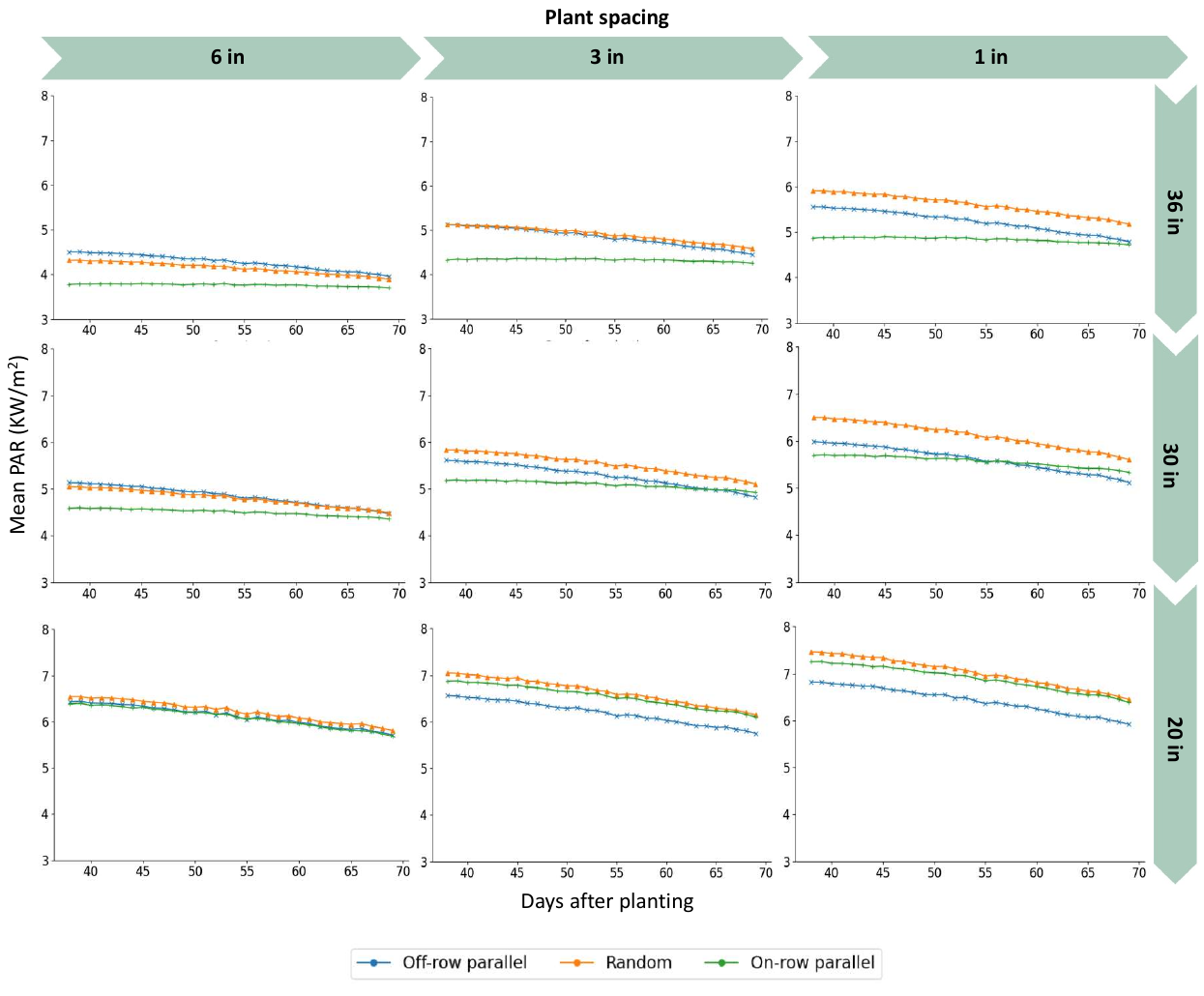}
    \caption{PAR interception for varying plant and row spacing for a typical growing season for Bismarck, ND.}
    \label{fig:bismarck}
\end{figure}

\begin{figure}[h!]
    \centering
    \includegraphics[width=\linewidth, height=0.65\textheight, trim={0mm 105mm 5mm 0mm},clip]{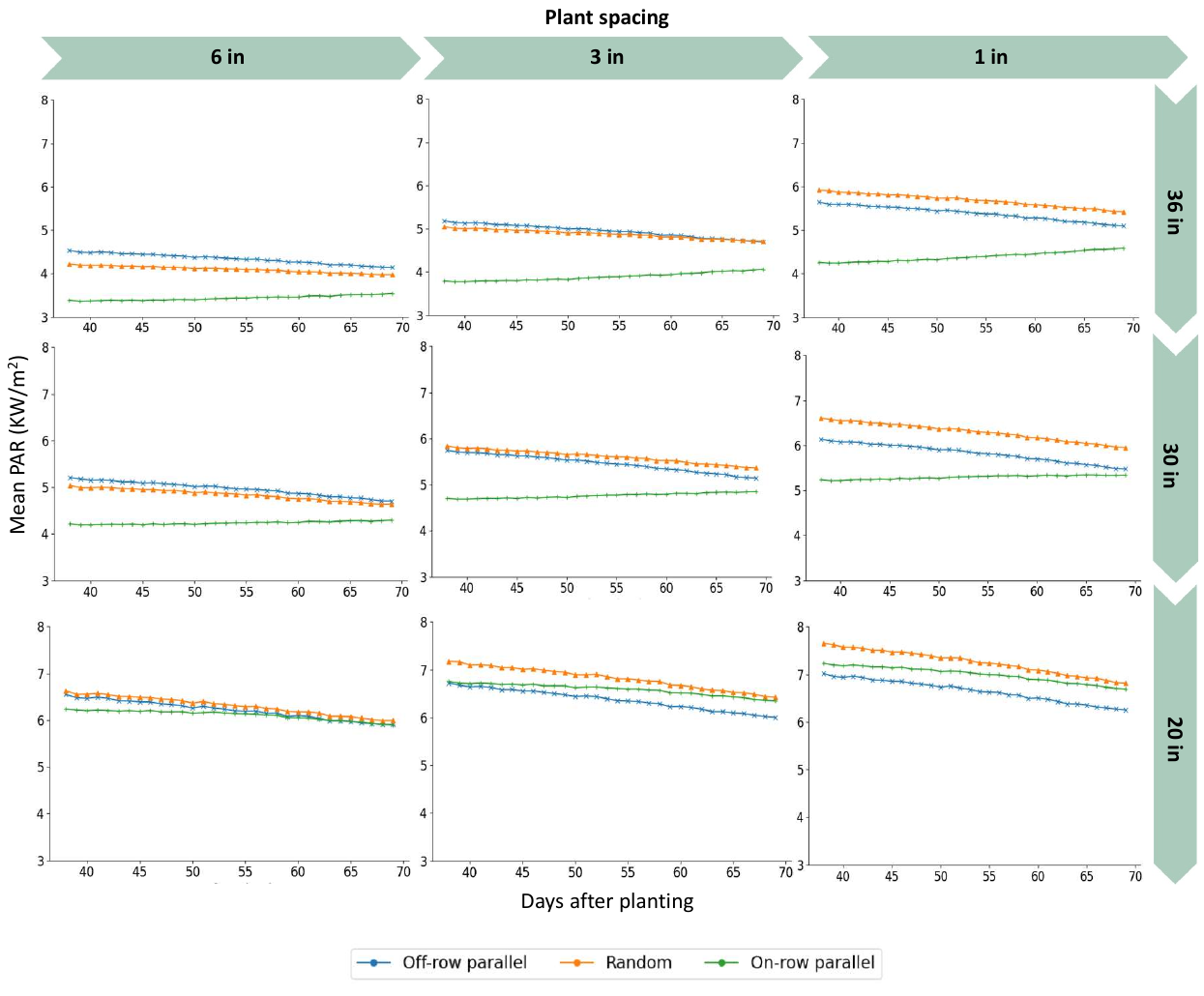}
    \caption{PAR interception for varying plant and row spacing for a typical growing season for Thomas county, KS.}
    \label{fig:thomasco}
\end{figure}

\end{document}